# Modeling the Experience of Emotion

Joost Broekens
Man-Machine Interaction group
Delft University of Technology

joost.broekens@gmail.com

Abstract

Affective computing has proven to be a viable field of research comprised of a large number of multidisciplinary researchers resulting in work that is widely published. The majority of this work consists of computational models of emotion recognition, computational modeling of causal factors of emotion and emotion expression through rendered and robotic faces. A smaller part is concerned with modeling the effects of emotion, formal modeling of cognitive appraisal theory and models of emergent emotions. Part of the motivation for affective computing as a field is to better understand emotional processes through computational modeling. One of the four major topics in affective computing is computers that *have* emotions (the others are recognizing, expressing and understanding emotions). A critical and neglected aspect of having emotions is the *experience* of emotion (Barrett, Mesquita, Ochsner, & Gross, 2007): what does the content of an emotional episode look like, how does this content change over time and when do we call the episode emotional. Few modeling efforts have these topics as primary focus. The launch of a journal on synthetic emotions should motivate research initiatives in this direction, and this research should have a measurable impact on emotion research in psychology. I show that a good way to do so is to investigate the psychological core of what an emotion is: an experience. I present ideas on how the experience of emotion could be modeled and provide evidence that several computational models of emotion are already addressing the issue.

Introduction

Computational models of natural phenomena are useful for two main reasons. First, the model itself can be used to simulate and predict the phenomenon that is modeled. Consider for example the weather. A detailed computational model of clouds, temperature fronts, pressure systems and geological factors predicts the weather for the next couple of days. This is obviously useful, and it is an inherent quality of the model. Second, any model is the instantiation of a theory or set of hypotheses, whether these hypotheses are simple or complex, widely validated or new. For a model to be computational, it needs to be executable by a computer (the model must be a computer program that can *run*). Regardless of the particular peculiarities of the computer system that is used to run the program, a fairly detailed description of the model is always needed. The computer needs detailed step-by-step instructions that match the model. Therefore, a *computational model* is a detailed instantiation of a theory or set of hypotheses. The predictions produced by a running computational model are predictions of the theory that model is based on. Obviously, the usefulness of these predictions critically depends on two factors: the credibility of the theory used as basis for the model, and the credibility of the extra assumptions that were needed to build the computational instantiation of that model (the correctness of the implementation). This means that a computational model also has a non-inherent, derived quality: it can be used to evaluate a theory. For example, predictions produced by a computational model of the weather can be used to evaluate the theory of the weather that underlies the computational model. Incorrect predictions motivate changes to the theory.

This view of computational models is not different in the area of affective computing. Affective computing is a research field that is concerned with computational modeling of emotion. Emotion in this field is interpreted in a broad sense. It is related to emotion recognition, emotion elicitation (production), emotion experience (feeling), and emotion effects (e.g., on cognition, behavior). I will discuss (the lack of) a definition of emotion later, but first I will give a more concrete example of a computational model of emotion that has both qualities, the inherent quality any model has that produces useful output, and the derived quality a model has provided that it is grounded in theory. Consider the work by (Gratch & Marsella, 2001). They propose a model of emotion elicitation, implemented in a pedagogical software agent. The role of the software agent is to guide a trainee through training. The software agent is used in virtual reality-based training, and can – partly as a result of the model of emotion -- deliver a more believable training simulation (more believable than training without emotional agents). This hopefully results in a training session that better matches reality, giving participants in the training a better preparation for real life situations. This is an inherent quality of the computational model of emotion. If the modelers did a faithful job of transforming the theory they used as basis for their model into a computational instantiation, which they did (Gratch & Marsella, 2004; Marsella & Gratch, 2009), then the behaviors of the agent are in essence predictions of the theory underlying the computational model. This is the indirect quality of the computational model of emotion. Predictions of this kind can be used as a motivation for changes to the emotion theory underlying the model (Broekens, DeGroot, & Kosters, 2008).

The inherent quality of a computational model is solely based on its potential for generating useful output. Useful means that the model serves some goal. In virtual training systems (e.g., (Gratch &

Marsella, 2004; Henninger, Jones, & Chown, 2003)) the goal is to enhance believability of the agent in order to increase the effect of the training on the trainees. In tutor and support systems (Bickmore & Picard, 2005; Graesser, Chipman, Haynes, & Olney, 2005; Heylen, Nijholt, Akker, & Vissers, 2003), the goal is to enhance the relation between the user and the system such that the system is used more often, longer and/or more effectively. In entertainment computing (Nakatsu, Rauterberg, & Vorderer, 2005), the goal is to increase enjoyment by increasing active experience (sufficient motor behavior of the user while playing for example a game) and presence (the feeling of being "in the game"). Affective computing techniques can be used to develop affectively interesting gaming characters (Broekens & DeGroot, 2004) but also to sense the user's affective state to adapt gameplay (Hudlicka, 2008). In autonomous agent research (Coddington & Luck, 2003; Gmytrasiewicz & Lisetti, 2000; Steunebrink, Dastani, & Meyer, 2008), the goal is to enhance (and reason about) the artificial agent's decision making process. It is obvious that if the reason for developing a computational model of emotion is different depending on the domain, then the usefulness of the model depends on the domain.

The derived quality of the computational model (i.e., its potential to evaluate the underlying theory) however does not depend on the domain. As mentioned above, this quality depends on the plausibility of the theory on which the model is based and the plausibility of the assumptions used to develop the computational instantiation of that theory. As a result, any model of emotion that fulfills these two criteria has derived quality, and can be used to evaluate the theory on which it was based.

Figure 1 depicts the two senses of usefulness just discussed. I explain the figure in detail. In psychology there are several types of theories of emotion that are suitable as basis for computational models of emotions (e.g., cognitive appraisal theory, factor-based theories of emotion, and motivation/drive/needs theories of emotion). These theories are widely cited and actively used as basis for current computational models of emotion. The computational models are developed for different domains (as explained above). The results of the studies are therefore domain-specific results. In affective computing, mostly the domain-specific results are reported upon. Such results include *our virtual character is more believable*, *our agent can reason more efficiently*, and *our tutor agent helps users to better learn*. These are good results, and a proportion of these results is fed back to the domain itself; the results are generalized. Such generalizations could include *emotional game agent (NPCs) are more enjoyable than non-emotional agents and we can explain that because they increase the feeling of presence*. And, *affective virtual characters increase the training effect of virtual reality training and we can explain that because the model of emotion increases believability of the agent and therefore the training is more effective*. Notice, however, that these generalizations are still domain specific and do not feed back to the emotion theory that was used as basis for the computational model of emotion. There is surprisingly little feedback (thin arrow from *Theory specific usefulness* to *Psychology*) to the psychology community on the predictions of their own theories generated by computational modeling (and I will substantiate this claim later). I call these results predictions on purpose, they are as much predictions of a particular emotion theory as a weather simulation is a prediction of a theory of the weather and a simulation of a super-nova explosion is a prediction of a theory of star formation and death. This lack of feedback to psychology is striking, given the fact that there definitely are affective computing researchers (e.g., (Breazeal & Scassellati, 2000; Broekens, 2007; Canamero, 2000; Gratch &

Marsella, 2004; Hudlicka, 2005; Lahnstein, 2005; Sloman, 2001), to name just a few), that try to faithfully model emotion according to the psychological theories and report the results relevant to the theory (denoted by the significantly thicker arrow from *Computer Science* to *Theory specific usefulness*).

In this position paper I argue that efforts in computational modeling of emotion should focus more on returning results to psychology, and I propose a research direction that can achieve this.

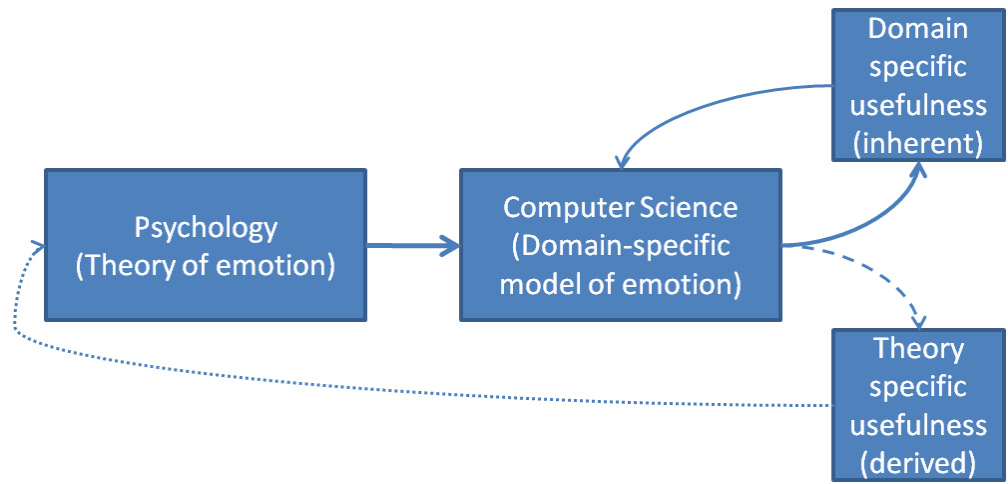

*Figure 1. Current state of the flow of results in affective computing research. Arrows denote the size of the impact of the results.*

Affective Computing Background and Future Challenge

In the last decade, affective computing has proven to be a viable field of research comprised of a large number of multidisciplinary researchers resulting in work that is widely published and used. Although this is not a review article, I feel it is appropriate to discuss the vast terrain affective computing by now covers. The reason for doing so is (a) to give an overview to those readers not familiar with the large amount of work that has been done, and (b) to show there really is a gap in the research focus.

First, a large body of research studies automatic emotion recognition in a wide variety of domains (Cowie et al., 2001; Hanjalic & Li-Qun, 2005; Pantic & Rothkrantz, 2000; Picard, 1997). These models typically try to affectively label facial expressions, sounds and movement. The main idea is to extract affectively relevant features from the signal, where the signal can be images, movies, sounds and gestures. Combinations of features together correlate with particular affective labels such as happy or sad (often as indicated by human subjects analyzing the same signals). Newer research tries to not only extract static features (such as the mouth having a certain angle) but also dynamic features, i.e., movement. These more detailed analyses are needed to differentiate between fake and genuine emotions. This branch of research is strongly related to signal processing and machine learning: try to identify relevant features, build a predictive model based on those features and use the model to label the signal.

Second, a large part of the affective computing research community is involved in computational modeling of causal factors of emotion in order to use these models in human-computer and human-robot interaction (for an overview see, e.g., (Bickmore & Picard, 2005; Breazeal, 2003; Fong, Nourbakhsh, & Dautenhahn, 2003; Hudlicka, 2003; Paiva, 2000)). Two aspects play a key role: modeling the emotion elicitation and modeling the emotion expression. One often does not go without the other (how to express an emotion without having a model that generates one, and why generating one if you don't express it). Important examples of these approaches are Kismet the social robot (Breazeal & Scassellati, 2000), companion robots such as the Aibo, the iCat, the Huggable and Paro (for review see (Broekens, Heerink, & Rosendal, in press; Fong et al., 2003)), and a large number of emotional agents used for tutoring systems (e.g.,(Graesser et al., 2005; Rickel & Johnson, 1997)) and virtual reality training (e.g., (Elliott, Rickel, & Lester, 1999)). Early research was primarily aimed at exploring the possibility of creating such affective agents (with the underlying assumption that affective agents are more believable than non-affective agents), while later research is more focused on measuring the effects on users of embedding such agents in particular domains (gaming, training, tutoring) and the effects on users of social and companion robots

Third, a smaller part of the affective computing community is explicitly concerned with modeling the effects of emotion, such as affective influences on cognition (Broekens, Kosters, & Verbeek, 2007; Canamero, 2000; Gadanho, 2003; Hudlicka, 2005; Marinier Iii, Laird, & Lewis, 2009; Velasquez, 1998), formal modeling of cognitive appraisal theory (Broekens et al., 2008; Gmytrasiewicz & Lisetti, 2000; Marsella & Gratch, 2009; Meyer, 2006), integrating emotional states in agent reasoning (Coddington & Luck, 2003; Meyer, 2006; Steunebrink et al., 2008) and models of emergent emotions, such as emerging from the interaction between a simple adaptive agent and its environment (Canamero, 2000; Lahnstein, 2005; Velasquez, 1998). Later in this article, I will come back to these modeling efforts, as some of these are surprisingly close to the research efforts that, according to emotion psychologists, are needed to study the experience of emotion (Barrett et al., 2007).

Two recent developments related to affective computing are social robotics (for review see (Fong et al., 2003)) and long-term human-computer relationships (see, e.g.,(Bickmore & Picard, 2005)). Key to these areas of research is the realization that, although computational modeling of emotion is needed for social robots and interfaces, it is not sufficient. Issues such as attachment and friendship, mutual dependency and shared understanding are essential too. Interestingly, (Breazeal & Scassellati, 2000) already started to study these issues with Kismet and (Dautenhahn, 1995) already argued these to be important for social intelligence in artifacts.

A general pattern in the affective computing research just reviewed is a strong focus on practical outcome. For example, a tutor agent, a virtual character, an affective interface, an emotion recognition mechanism, enhanced human computer interaction. As a consequence, many reported results relate to the inherent quality of the computational model of emotion: its ability to produce useful output in the domain for which it was developed. This obviously skews the number of results towards the "inherent" box instead of the "derived" box (Figure 1).

However, part of the motivation for affective computing as a field is the promise to better understand emotional processes through computational modeling. The question is what it means to "better understand emotion"? Or, more appropriately for the current discussion, when is a "better understanding" happening? I think that this must involve feeding back the results from computational studies of emotion to psychology. Considering the lack of citations *to* affective computing *from* leading publications on emotion, such as (Barrett et al., 2007; Frijda, Manstead, & Bem, 2000; Scherer, Schorr, & Johnstone, 2001), and more importantly *The Handbook of Emotions* (M. Lewis, Haviland-Jones, & Barrett, 2008) the affective computing community did not add a lot to the understanding of emotion, at least not to the extent that psychologists cite the work.

This is rather counterintuitive as in psychology there is a strong need to better understand the processes and mechanisms that underlie emotion. Sometimes mathematical and computational principles are even used to do this (M. D. Lewis, 2005; Reisenzein, 2009; Wehrle & Scherer, 2001). Also approaches that are less mathematically inclined do express a need for a better understanding of the underlying mechanisms of emotion, such as in (Barrett et al., 2007), but these needs are expressed in terms of neural workings. Neural workings, though, can still be simulated to a certain extend. The point is that to understand mechanisms, simulation is a very powerful research method. A simulation of a phenomenon not only shows a relation between constructs but also posits a possible explanation for why that relation exists.

In her seminal book and in later papers Picard (Picard, 1997, 2003) and others state that one of the four major topics in affective computing is computers that *have* emotions (the others are recognizing, expressing and understanding emotions). In psychology, a critical and neglected aspect of having emotions – of emotion in general – is the *experience* of emotion (Barrett et al., 2007). The experience of emotion can be summarized as the answer to the following four questions: what does the phenomenological content of an emotional episode look like, how does this content change over time, when do we call the episode emotional, and how do neurobiological processes instantiate the experience? Surprisingly little computational modeling effort in the affective computing community has these topics as *primary focus*. I think that a focus on these topics helps to restore the impact balance because it directly contributes to answering questions relevant to emotion psychology.

The launch of a journal on synthetic emotions should spark research initiatives in *theoretical affective computing* (or *computational affective science* if you like that term better). This research should be cited in the psychology literature. Currently, this does not happen often, while the inverse citation relation is abundant: all affective computing researchers know it is practically impossible to get a paper published without referring to the standard emotion works.

In this article I argue that the best way to restore the impact balance is to investigate the psychological core of what an emotion is: *an emotional experience*. I present ideas on how the experience of emotion could be modeled and what to do with such a model. More importantly, I provide evidence that – from a psychological point of view – some computational models of emotion are already addressing the issue. I therefore propose a challenge: to have one chapter on computational modeling of emotion in the fourth edition of *The Handbook of Emotions*.

## Not a definition of emotion

Emotion is a complex topic, and agreement on one solid definition does not really exist. I will not attempt to define emotion here, as many excellent works have been published from different perspectives that together do much more credit to the diverse and multimodal nature of emotion (Frijda et al., 2000; LeDoux, 1996; M. Lewis et al., 2008; Ortony, Clore, & Collins, 1988; Panksepp, 1998; Picard, 1997; Rolls, 2000; Scherer et al., 2001). In this section I explain what the different emotion-related terms usually refer to, and the above-mentioned references are a collective source for this explanation.

Typically, affect refers to the underlying core of emotion, mood and affective attitude towards persons and things. Emotion, mood and affective attitude are different but strongly related and influence each other. In general, emotion is related to facial expression, feeling, cognitive processing, physiological change and action readiness. Furthermore, emotion refers to a short but intense episode that, in addition to the previously mentioned aspects such as facial expressions, is characterized by "attributed affect to a causal factor". An emotion is a noticeable if not powerful experience. For example, I feel (and notice I am) happy about seeing an old friend. In contrast, mood refers to a silent presence of moderate levels of affect. I can feel frustrated for half a day without knowing why. Mood is not (consciously) attributed to a causal factor. Affective attitude refers to how one generally feels about something or someone, not specifically because of that thing or person. For example, I *like* popular science books, and I feel *enthusiastic* about theme parks. To complicate matters a little, affect is also used as commonplace term for everything that has to do with the above.

There are several theoretical views on how to think about emotion. These views can be categorized in multiple ways, but I find the following categorization that uses two axes particularly useful. The first axis defines the level of abstraction at which emotion is studied: social, psychological, biological, and physiological. The second axis defines the way emotion is represented: categories of emotion, components that form an emotion, and principle factors. For example, the well-known six basic emotions as proposed by Paul Ekman are categorical (fear, anger, happiness, etc.). Cognitive appraisal theories are componential, as these describe emotion as a combination of the activation of different sub processes (evaluation of an event in terms of novelty, goal conduciveness, etc.). On the other hand, (Russell, 2003) proposes a description of emotion using two continuous factors (Pleasure, Arousal).

Disregarding these different views, many (if not all) emotion researchers believe that there are two common affective factors that are useful to describe a mood, emotion or attitude: valence and arousal. The difference in opinion is not so much about these factors but about how to interpret what they are. Are these factors the emotion, do they represent something real in the brain and if so which brain areas are involved, are they independent (orthogonal), are they artifacts of statistical analysis of many factors, etc. The same holds for theories that explain emotion elicitation (where does the emotion come from). Although most psychologists agree that an emotion is the result of an evaluation of the situation in terms of personal relevance, the argument is about what the evaluation looks like (is it cognitive, conscious, holistic, component-based, automatic, hard-wired, social, etc.). For a recent overview and reflection upon emotion and emotion elicitation, see (Gratch, Marsella, & Petta, 2009). For a quick and broad introduction to the different emotion theories and the history of these see chapter 5 in (Eysenck, 2004) or chapter 3 in (LeDoux, 1996).

The citation asymmetry and the experience of emotion

I first substantiate the claim that there is an impact asymmetry between affective computing and the psychology of emotion, and I will do this in a very simple way. If the latest edition of the standard psychological reference manual for emotion research scholars (M. Lewis et al., 2008) does not contain even a single reference to the most important results in affective computing, while all affective computing literature contains loads of references to emotion theories, then there is a serious asymmetry. I would claim there is no such thing as affective computing for beginning scholars in emotion research.

Why is the impact of affective computing on psychology so small? I argue there are three main causes that might explain this asymmetry in citation. The first one is that the affective computing community has a strong focus on inherent model quality, i.e., the ability to use the model in a particular domain is seen as the end result of a study (thick arrow going to the "inherent" box in Figure 1). This is clear from the literature overview just presented: many approaches use emotions for a particular purpose in a system, and when that has been done successfully they consider the study done. Computer scientists doing such studies do not relate their results to the theory of emotion they used. This is not needed. If the goal of your research is to, e.g., develop an affective interface, the results are evaluated in terms of the effectiveness of the interface, not in terms of the predictions of the theory used as basis for the interface.

The second cause has to do with how results are communicated to emotion psychologists, assuming there are any (how are the results depicted by the thin arrow to the "psychology" box in figure 1 communicated). Research results are not communicated in an effective way to the psychological research community. Computer scientists have a different language and have different publication outlets. A related issue is that in computer science it is common to publish results on conferences, and many of these conferences are not indexed by popular psychological search engines. Even when the results are related to the emotion theory, these are often described in too technical a manner (including formalisms, program code, etc.) or use unconventional terminology. A related aspect is that emotion psychologists are used to read lengthy discussions of how a particular new finding relates to findings and theories of others, while computer scientists do not discuss their results in this way. Computer science research is problem oriented, while psychology research is exploratory. As a result, when the problem is solved (e.g., I have an emotional avatar) the research is done.

Third, assuming emotion psychologists find and understand affective computing publications, they do not feel that the studies add to their understanding of emotion. I personally think this is much more serious an issue than the first two, as it means there actually isn't a "derived" quality box (Figure 1). This feeling can be due to two things. The implicit assumptions needed to develop a computational model based on a theory of emotion cloud the validity of the predictions coming out of the running model. As a result it is difficult to claim that the computational model is a valid instantiation of the theory. Second, the topics addressed in affective computing are simply not relevant to the understanding of emotion. The first issue is a methodological one (Broekens et al., 2008), and although important I will not address here. The second issue is the one I will discuss in the rest of this article.

Do affective computing researchers address the relevant topics from the point of view of emotion psychologists? I think the answer to this question is sometimes. Consider emotion recognition research. The typical problem that is addressed is "how can we extract the (real vs. fake) affective meaning of an expression"? This question is seen as the problem, and the tools used to solve this problem are signal processing and machine learning. As a result, advances in this field are mainly advances in signal processing and machine learning, not in understanding emotion expression. Obviously the computer science model builders understand more and more of emotion expression, but they use insights from the psychological literature to do so. Automatic emotion recognition is a special form of signal processing (typical publication venues are the signal processing journals and conferences). The same argument can be made for artificial emotion expression. The problem to be solved is "plausible emotional expression", and as a result, expressing emotions is a particular form of rendering (typical publication venues are the graphically oriented and robotic journals and conferences).

So, when do affective computing researchers address relevant problems for emotion psychologists? I think there are two areas that are directly relevant to emotion psychologists: computational models of emotion elicitation (production) and computational models of emotion effects on thought and behavior. These areas directly touch the heart of the questions surrounding *emotion experience*: what does the phenomenological content of an emotional episode look like, how does this content change over time, when do we call the episode emotional, and how do neurobiological processes instantiate the experience?

In the remainder of this article I argue that affective computing modeling efforts can help our understanding of the experience of emotion. I also present evidence that affective computing researchers are already doing this, thereby supporting the feasibility of modeling the experience of emotion.

## Modeling the experience of emotion

In a recent review of emotion research it has been argued (successfully in my opinion) that what is currently lacking is a focus on what an emotion *is* (Barrett et al., 2007). According to the authors, this is due to a couple of reasons (including a focus on behaviorism, fear of phenomenologically oriented research, etc.) that I will not further discuss here. However, an important observation is that there has been a strong focus on causal (and preferably external) factors for emotion *not* on a thorough description of what the experience of emotion feels like and how such feelings arise (Barrett et al., 2007). The authors argue for efforts into investigating both content, a phenomenological description of the experience itself, and process, an explanatory description of the relation between the phenomenological experience and the biological and neurological underpinnings. In essence the authors advocate parallel study of both the "hard" problem of emotion (phenomenological structure and links to neurobiological mechanisms) and the "easy" problem of emotion (biological, social and psychological functions, antecedents and consequences) (Chalmers, 1995). The authors' motivation for this approach is a good one: one does not know what to explain if there is no adequate description of the explanandum, i.e., the experience of emotion. Following (Barrett et al., 2007), a phenomenological description of emotion is a description that includes "affect, perceptions of meaning in the world, and conceptual knowledge about emotion bound together at a moment in time, producing an intentional state where affect is experienced as having been caused by some object or situation". Such a description contains the following key factors.

- *Affective associations*: such as past feelings, hypothetical feelings, and online experiences.

- *Core affect*: a description of the pleasure and arousal intensities during the emotive period.

- *Arousal content*: a feeling of being active versus passive, including the difference between actual and felt arousal, relation to attention processes, etc.

- *Relational content*: the dominance relation between people, or more generically, the social context of an emotion.

- *Situational content*: the appraisal (in the cognitive appraisal theory sense) content including causal events, goal conduciveness, novelty, and norm compatibility.

- *Appraisal detail*: the commonalities and subtle differences between different forms of anger, sadness, etc. (Barrett et al., 2007) call this part "beyond appraisal dimensions". An important distinction made is the difference between felt and reported experiences. Currently many of the appraisal theories address the latter not the first. Both are important, but a good description of the felt experience is important from a phenomenological point of view (and more useful when searching for neural correlates).

An important related issue is the relation between time and the experience of emotion. What is the dynamic interplay between the above mentioned aspects, and how do they vary over the onset and decay of an emotional episode?

As mentioned above, there are two ways via which the experience of emotion could be unraveled (Barrett et al., 2007). First we should have a good description of the phenomenological content of emotion experience, and second we should investigate the neural correlates of this content (what the authors call a neural reference space).

There are *three* ways in which computational modeling can help. First, to investigate the neural correlates of emotion experience, computational models can be used to simulate biologically plausible neural networks involved in emotion, as is commonly being done in computational cognitive neuroscience. This type of model is strongly grounded in biological and physiological accounts of emotion but the findings generated are difficult to link to the phenomenological content of the experience of emotion. The conceptual gap between physiological and neurological theories of emotion that are suitable for computational modeling on the one hand, and a phenomenological description of emotion on the other, is too big.

Second, we can use formal modeling techniques to describe the emotion experience itself. The model is a formal representation of what is known about the content and flow of the experience of emotion, but cannot be used to generate new predictions, because the mechanisms responsible for the content and flow are not modeled. For example, one could model the phenomenological content of emotion using a network of nodes probabilistically connected to each other with edges. Each node contains a content description, and the edges define possible transitions. As a result, this network specifies the flow of emotion experience. The network can be executed to simulate possible flows of phenomenological experience, but it can never generate new findings.

Third, computational modeling can help to find plausible mechanisms that explain and predict the phenomenological content of emotion. In this case, the model consists of mechanisms specified in a language positioned between neuronal processes and phenomenology predicting phenomenological content. In this article I focus on this type of computational models. The simulation results produced by these models are interpretable in terms of the experience of emotion, while at the same time the model proposes mechanisms that could be responsible for the generation of the experience. In other words, affective computing can help emotion psychology by developing *generative phenomenological models* (for want of a better term).

I present three examples of existing generative phenomenological models. Two key characteristics of such models are (a) that the model is based on a theory that explains emotion elicitation (such as appraisal theories) and (b) the results of the predictions are interpreted in terms of a phenomenological description of emotion experience. The advantages of such an approach are that the model can be executed, can be used to generate new experiences, and these experiences can then be verified in (or compared to) psychological experiments. The three different approaches discussed are a logic, formal approach by (Steunebrink, Dastani, & Meyer, 2007), an agent-based approach by (Marsella & Gratch, 2009), both of which are cognitive appraisal based, and a behavioral, biological approach by (Lahnstein, 2005) that is compatible with an emotion theory such as the one proposed by Rolls (2000). Each study sheds light on the six previously mentioned aspects of a phenomenal description as proposed in (Barrett et al., 2007), by explaining or predicting parts of the content of emotion experience.

(Lahnstein, 2005) proposes a model of the onset and decay of an emotive episode as a result of an anticipatory and subsequent reactive evaluation phase. In this model she uses a simple robot that is able to learn behavior based on reward, and control its behavior based on the prediction of future reward. The model uses a form of reinforcement learning. In the discussion, we assume the model has had several learning experiences, i.e., the model is partly trained. She proposes that the valence part of the emotive episode can be modeled as a combination of the expected reward -- the reward that is anticipated given a certain action --, and the experienced reward – the reward received when the action is executed. According to (Lahnstein, 2005), the positive valence signal has four typical dynamics, three of which I think are particularly insightful (see Figure 2). Importantly, she proposes phenomenological qualities for these different dynamics. In the first case, the prediction of positive reward would correlate with feelings of hope, optimism and positive expectancy (phase 1) and the evaluation of the received reward that is smaller than expected would correlate with disappointment (phase 2). In the second case, the experience of phase one is the same, while the experience of phase two would correlate with happiness, contentment and satisfaction. In the third case, the first phase is again the same, but the second phase is correlated with happiness (and I would add positive surprise as the reward is larger than expected). I think we can easily add to her proposal plausible phenomenological accounts of these three dynamics if the valence signal is negative (e.g., fear in the first phase, and in the second phase panic when the punishment is even worse, relief when the punishment is less, and despair when the punishment is confirmed). In a series of small experiments she shows that these positive valence dynamics actually occur in a simple learning robot. By doing so, she predicts, using a generative model, how the content of the emotion experience would look like for an adaptive organism with regards to *core affect*. Further, she also predicts what the online *affective associations* would be (happiness, disappointment, etc.).

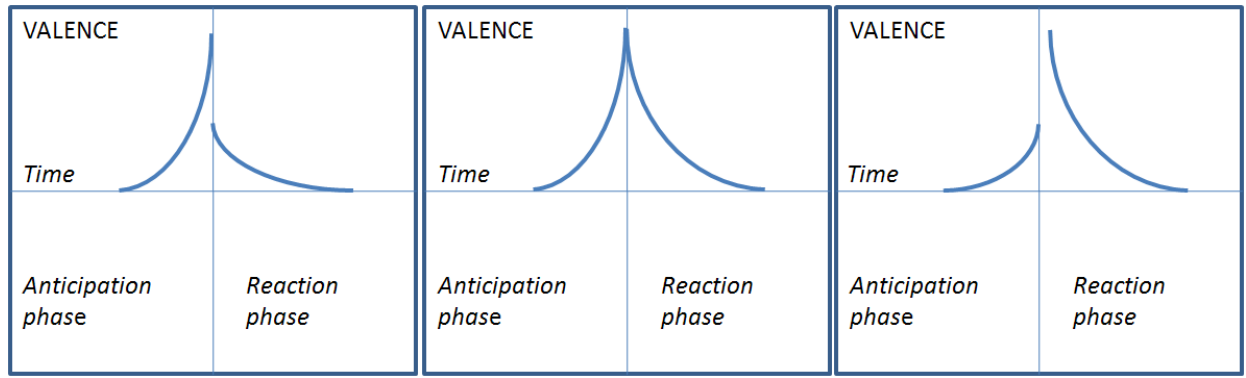

*Figure 2. Three different situations of the predicted and received reward, taken from Lahnstein (2005).*

The results just presented are compatible with the theory of emotion proposed by (Rolls, 2000), who states that emotion is the result of a combination of six reinforcement-related factors including whether reward or punishment is given or withheld, the intensity of the reinforcement and the occurrence of both a reward and a punishment at the same time. Rolls behavioral account of emotion is a significant (although debated) addition to our understanding of emotion in the context of adaptive behavior. I would claim that the modeling work by (Lahnstein, 2005) is a significant contribution to our understanding of the possible dynamics of emotion in relation to adaptive behavior.

A very different branch of study in affective computing is how to formalize appraisal theories. As an example, I will discuss a recent attempt by (Steunebrink et al., 2007). In their study they formalize the OCC appraisal model by (Ortony et al., 1988) by extending an already existing BDI (Belief, Desire, Intention) based agent programming logic. The details of this language are out of scope in this article, but I will have to explain a little about the underlying ideas. First, they have a system that is capable of formally representing beliefs, goals, agents, plans, abilities, commitment and the execution of plans as well as the relations between these elements. Second, they define the different emotions that can be found in the OCC model, for example, they define *hope$_i$(p, g)* as the emotion of hope of an agent *i*, that a particular plan *p* will end up satisfying goal *g*. In an analogue fashion they define *fear$_i$*(*p*, -*g*) and all other emotions in the OCC model. Third, they formalize what hope would be for an agent according to the OCC model given that the agent has a formal structure around events, goals, etc. as they defined it. So for hope this would be *hope$_i$*(*p*, *g*) ↔ (*I*(*p*, *g*), *Commitment*(*p*)), which means that an agent hopes for plan *p* to fulfill goal *g* if and only if it has the intention to fulfill goal *g* with plan *p* and it has a commitment to executing plan *p*. This, in their view, maps to what is describe in the OCC model about hope; hope is being pleased about the prospect of a desirable event. Prospect is thus interpreted as "having a plan for", and desirable is interpreted as "being a goal". Fear on the other hand is formalized as follows: *fear*(*p*, -*g*) ↔ (*hope*(*p*, *k*) & *Belief*(*execute*(*p*)-*g*), which in their formal system means that fear about not fulfilling a goal *g* due to a plan *p*, occurs if and only if there is a hope for that plan fulfilling the goal, and the agent beliefs there is a possibility that execution of that plan might end up in not fulfilling the goal. In other words, fear involves uncertainty about successfully executing a plan aimed at a particular goal.

Regardless of what one thinks of the different assumptions that are the basis of this formalization (e.g., it might be argued that a prospect does not necessarily imply having a plan), the formal notation does serve emotion research, and in particular the *situational content*. First it makes very explicit what the consequences are of particular assumptions and theory translations. For example, fear in their case can only be derived from hope, meaning that the experience of fear would always be accompanied by a feeling of hope. This is a testable prediction, first to be evaluated on face value by emotion psychologists, then by experimentation. Second, one can derive things about the situation that elicits the emotion. A striking one is the following derivation. If an agent does not believe in the existence of a plan to accomplish a goal, it will neither hope for that goal nor fear the consequences (Steunebrink et al., 2007). This can be logically derived in a very easy way, as having a plan is a precondition for having hope and having hope is precondition for having fear. Third, formalization might learn us new things or confirm things we knew but never made explicit. For example, fear and hope differ by the fact that fear involves uncertainty about the plan. This can be derived from the following: (a) the intention for a particular plan is a prerequisite for having hope, (b) this intention exists if and only if the agent believes it can execute the plan, believes it wants the goal and believes that executing the plan actually results in the goal (Steunebrink et al., 2007), and (c) fear exists if and only if there is hope and the agent believes that the execution of the plan might *not* result in the goal. As a consequence, the agent believes that the execution of the plan could result in both the goal and not the goal. This is the essence of uncertainty. Again, this is a prediction entirely derived from a formal structure based on a theory of emotion (OCC). I claim that these kinds of models and their predictions are an important contribution to the phenomenological description of appraisal: the model describes how the *situational content* influences the resulting emotion. Unfortunately, the way the results are reported prevents the average emotion psychologist to pick up on these results, apart from perhaps those with a particular interest in formalisms (Reisenzein, 2009).

Finally, I will discuss an example of how affective computing can result in theory refinement useful for understanding the experience of emotion. (Marsella & Gratch, 2009) propose a computational model of appraisal, called EMA (EMotion and Adaption), that can be executed and used to explain the dynamics of an emotional episode. This is an important aspect of the experience of emotion: how does the emotion unfold over time, and what are the detailed specifics of the emotion. I will summarize the approach taken by Marsella and Gratch as well as their own example, to demonstrate that their model can indeed shed new light on emotion dynamics. The main assumption in their model is that appraisal processes (the processes that evaluate the relation between an agent and its environment) are always quick and shallow. Now, how can such quick and shallow processes result in complex emotions such as blame and guild, classically associated with effortful cognitive processing? The mechanism they propose is cyclic re-appraisal. Appraisal processes continuously evaluate external and internal events, and in parallel cognitive processes infer meaning and predict longer term consequences. New information generated by both types of processes as well as new events occurring in the world are evaluated continuously, and therefore the appraisal processes continuously reevaluate the agent-environment state. Because the appraisal processes are always shallow, but the state can become more complex (e.g., a remembered or a cognitively inferred future state), complex emotions occur later as a result of the quick and shallow evaluation of this more complex state (relief occurs after a couple of cycles, while fear for a nearing object can result immediately). So in essence, the authors claim that the complexity of appraisal is not in the appraisal processes but in the state generation mechanisms, and that the "apparent complexity" of certain appraisal processes (such as blame, guild, etc.) is an emergent property of the cyclic re-appraisal process. This all would be limited to very interesting philosophical contemplation, if it were not for the fact that EMA is not only a theory but also a computational model that can be used to verify these hypotheses. They show how the case of a bird that unexpectedly flies into the window during a lab experiment involving actors can be modeled accurately with their computational model. I will only summarize their main findings, by explaining the two key phases in the re-appraisal process of the bird scenario. In the first phase, the actor hears the sound of the bird flying into the window. This event triggers the appraisal processes, of which one responds (*expectedness*). It signals that this event is not expected, resulting in the emotion of surprise. As a result of the actions associated with surprise (attention-focusing, action-readiness) the actor sees a bird close by. This event (see_bird) triggers *expectedness* again, again resulting in surprise. The state of readiness and the nearing object cause the agent to infer that there might be injury of the actor involved. This fact (injury_possible) triggers the *desirability* appraisal process. The process evaluates the situation low desirable, and, to summarize a couple of steps here, the actor prepares to strike the bird and moves back to protect her head. The resulting emotion is anger. The second phase now kicks in. As a result of moving back the perspective towards the bird changes. New information is available (the bird is no longer approaching, it is a small bird). The potential for injury is reevaluated, and the situation is re-appraised as positive. As a result of the positive evaluation and the by now predictable situation, the emotion of relief is triggered. However, at the same time, the situation of the bird is now more clear and appraised as low desirable for the bird resulting in concern.

This new version of the EMA model is an interesting case for this article. It does exactly what would be expected from a computational model of emotion experience. It is thoroughly based on assumptions from the psychological literature. It proposes an executable mechanism that is used to explain observed emotional behavior of the actor as consequences of these assumptions. It focuses on *appraisal detail* and *situation content*, and proposes a detailed description of the content as well as the time course of the appraisal and correlated emotional experience. (Marsella & Gratch, 2009) clearly show that if the assumptions of a computational model are made explicit, the model can be seen as an executable instantiation of a theory. In this particular case, the authors opted for validating their own theory, but their approach can equally well be used for evaluating theories of others. As a matter of fact, it would be interesting to see predictions of observable human behavior generated by EMA.

These three examples show that appraisal detail, situation content and core affect can be computationally modeled to better understand the dynamics of the experience of emotion. All three models are based on emotion theory, and when executed generate testable predictions. Furthermore, the three models are insightful with respect to the mechanisms that could be responsible for the flow of emotion experience.

Conclusion

In this article I show that affective computing is not picked up in standard psychological emotion literature. I propose three main reasons for this. First, affective computing researchers often focus on the end result of a model of emotion (the models *inherent quality*), such as the ability to correctly recognize human emotion expression and the ability to produce plausible artificial emotions. The validity of the model they use to generate these results is not their primary concern. As a result this validity is not evaluated nor reported upon and the computational model cannot be judged on its *derived quality*, the ability to evaluate the theory it was based on. Second, the way affective computing researchers report their research is often not compatible with the way emotion psychologist are accustomed to read research results. Affective computing researchers could invest a little more effort in trying to discuss their results in relation to existing theories and findings in the emotion literature, something that is done, for example, in the work by (Marsella & Gratch, 2009), but not in the work by (Steunebrink et al., 2007). Both studies contain findings relevant for psychology, as I have argued in this article, however, the latter fails to elaborate on these findings in a for psychologists understandable way. Third, affective computing researchers typically attack problems that reside in the "technical" domain: emotion recognition, emotion expression rendering, virtual reality, gaming, etc., while emotion psychologists want to better understand emotion per se. I have shown that affective computing is able to help address what has been identified as the core of emotion research (Barrett et al., 2007), namely the experience of emotion. I also show how this can be done, and explain in detail three different approaches that do so. I think that these three examples show that affective computing is ready for a split. A long-standing promise is that affective computing can shed light on feelings and what it means to have emotions. I believe it can, but for that we need to be conscious about why a particular model of emotion is developed: is it outcome oriented or theoretical. This is the split; applied affective computing next to theoretical affective computing. If one does not like such an applied / theoretical distinction, then please note that these are just labels to emphasize the rational for doing the research and these approaches are by no means mutually exclusive. Alternatively, we could simply adopt the term computational affective science (with an eyewink to computational cognitive science).